\documentclass[letterpaper]{article} % DO NOT CHANGE THIS
\usepackage{aaai23}  % DO NOT CHANGE THIS
\usepackage{times}  % DO NOT CHANGE THIS
\usepackage{helvet}  % DO NOT CHANGE THIS
\usepackage{courier}  % DO NOT CHANGE THIS
\usepackage[hyphens]{url}  % DO NOT CHANGE THIS
\usepackage{graphicx} % DO NOT CHANGE THIS
\usepackage{natbib}  % DO NOT CHANGE THIS AND DO NOT ADD ANY OPTIONS TO IT
\usepackage{caption} % DO NOT CHANGE THIS AND DO NOT ADD ANY OPTIONS TO IT
\usepackage{algorithm}
\usepackage{algorithmic}

\usepackage{newfloat}
\usepackage{listings}
\DeclareCaptionStyle{ruled}{labelfont=normalfont,labelsep=colon,strut=off} % DO NOT CHANGE THIS
\floatstyle{ruled}
\newfloat{listing}{tb}{lst}{}
\floatname{listing}{Listing}
\title{Contrastive Multi-Task Dense Prediction}
\author {
    Siwei Yang\textsuperscript{\rm 1,2},
    \quad Hanrong Ye\textsuperscript{\rm 2},\quad
    Dan Xu\textsuperscript{\rm 2}
}
\affiliations {
    \textsuperscript{\rm 1} Tongji University\\
    \textsuperscript{\rm 2} Hong Kong University of Science and Technology\\
    swyang.ac@gmail.com, hyeae@cse.ust.hk, danxu@cse.ust.hk 
}
\usepackage{bibentry}
\usepackage{subfigure}
\usepackage{booktabs}
\usepackage{amsfonts}
\usepackage{amsmath}
\usepackage{MnSymbol}
\usepackage{multirow}
\newcommand*{\eg}{\emph{e.g.}}
\newcommand*{\ie}{\emph{i.e.}}
\def \etal{\emph{et al.}}
\DeclareMathOperator*{\argmax}{arg\,max}

\begin{document}

\maketitle

\begin{abstract}
This paper targets the problem of multi-task dense prediction which aims to achieve simultaneous learning and inference on a bunch of multiple dense prediction tasks in a single framework. A core objective in design is how to effectively model cross-task interactions to achieve a comprehensive improvement on different tasks based on their inherent complementarity and consistency. Existing works typically design extra expensive distillation modules to perform explicit interaction computations among different task-specific features in both training and inference, bringing difficulty in adaptation for different task sets, and reducing efficiency due to clearly increased size of multi-task models. In contrast, we introduce feature-wise contrastive consistency into modeling the cross-task interactions for multi-task dense prediction. 
We propose a novel multi-task contrastive regularization method based on the consistency to effectively boost the representation learning of the different sub-tasks, which
can also be easily generalized to different multi-task dense prediction frameworks, and costs no additional computation in the inference.
Extensive experiments on two challenging datasets (\ie~NYUD-v2 and Pascal-Context) clearly demonstrate the superiority of the proposed multi-task contrastive learning approach for dense predictions, establishing new state-of-the-art performances.
\end{abstract}

\vspace{-7pt}
\section{Introduction}\label{sec:intro}
%CV and DL are moving rapidly these years.
Dense prediction tasks such as semantic segmentation~\cite{zhao2017pyramid,chen2018encoder,song2019learnable}, depth estimation~\cite{xu2017multi,xu2018structured,cheng2019learning}, and saliency detection~\cite{wang2018detect,hou2017deeply} empowered with deep learning techniques are moving very rapidly in the recent years, and methods with convolutional neural networks (CNNs) have demonstrated great improvement on these different tasks.
However, most of the methods usually address these dense prediction tasks separately which results in low efficiency in real-life applications utilizing large-capacity CNNs. More importantly, simultaneous modeling of multiple different tasks allows us to capture the interior relationships and interactions among the different tasks, being able to realize a more powerful and higher-level perception system.

\begin{figure}[!t]
\centering

\subfigure[Triplet definiton w/ SemSeg labels]{
    \includegraphics[width=0.45\linewidth]{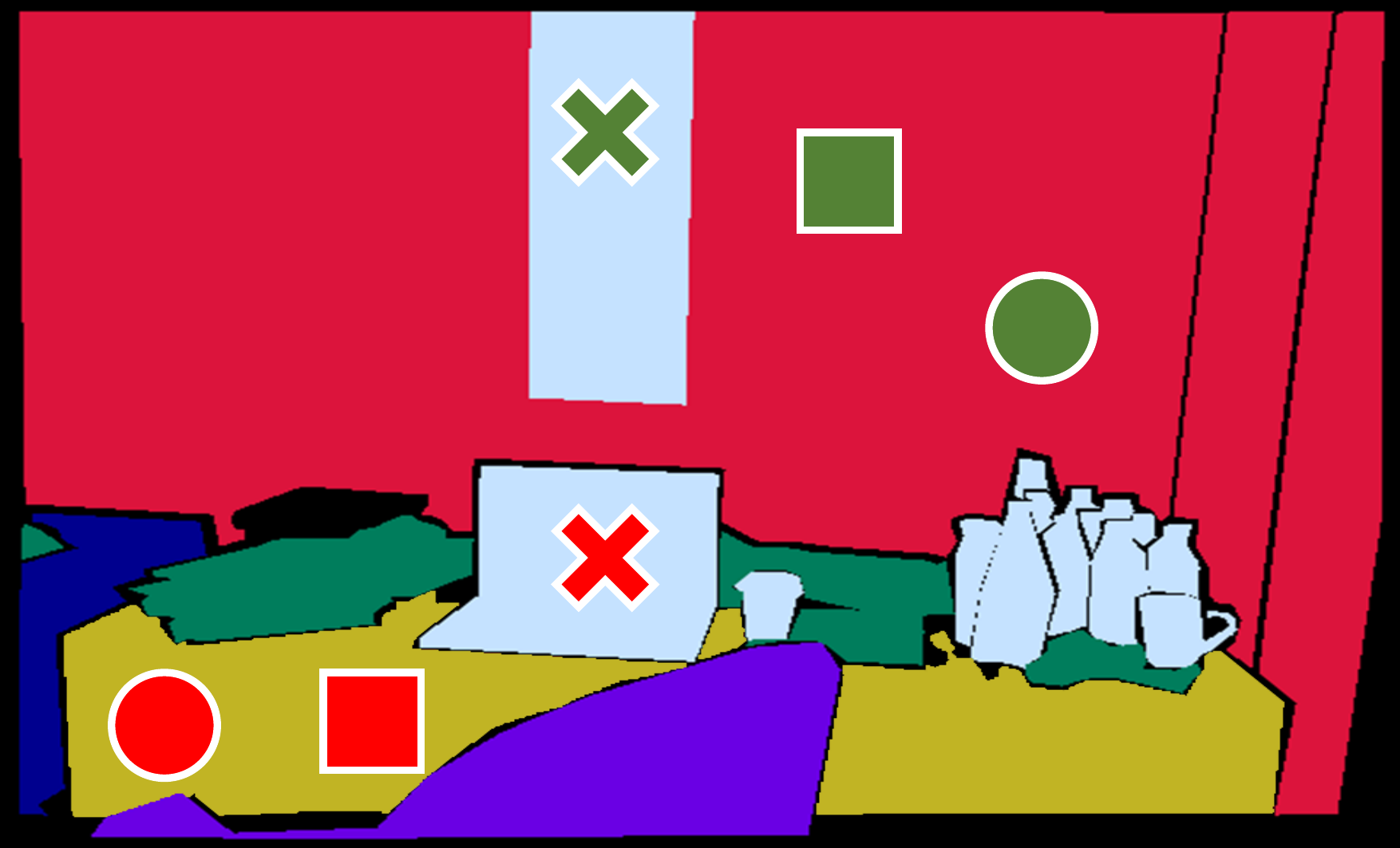}
}
\subfigure[Distance between SemSeg features w/ SemSeg labels]{
    \includegraphics[width=0.45\linewidth]{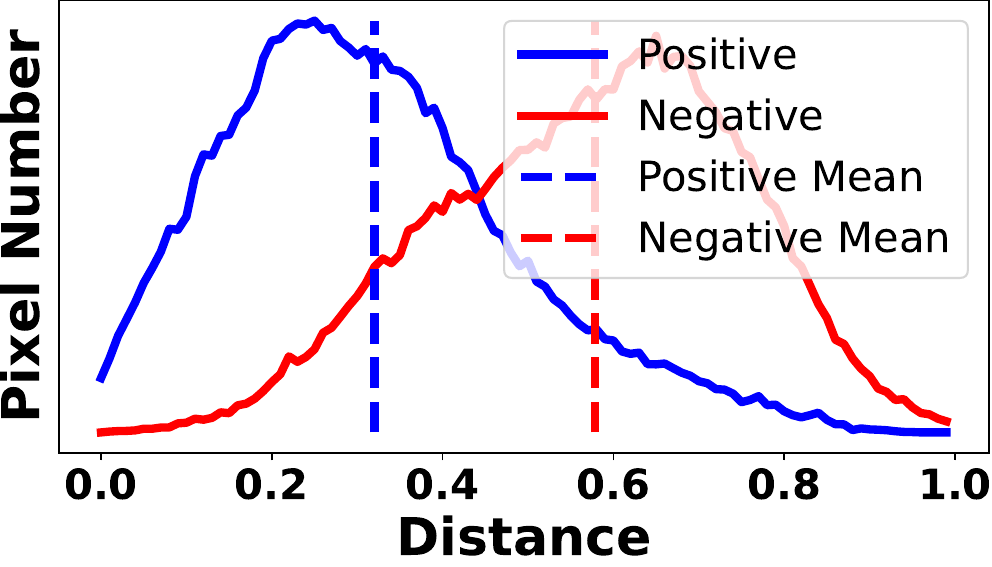}
}
\subfigure[Triplet definiton w/ Depth labels]{
    \includegraphics[width=0.45\linewidth]{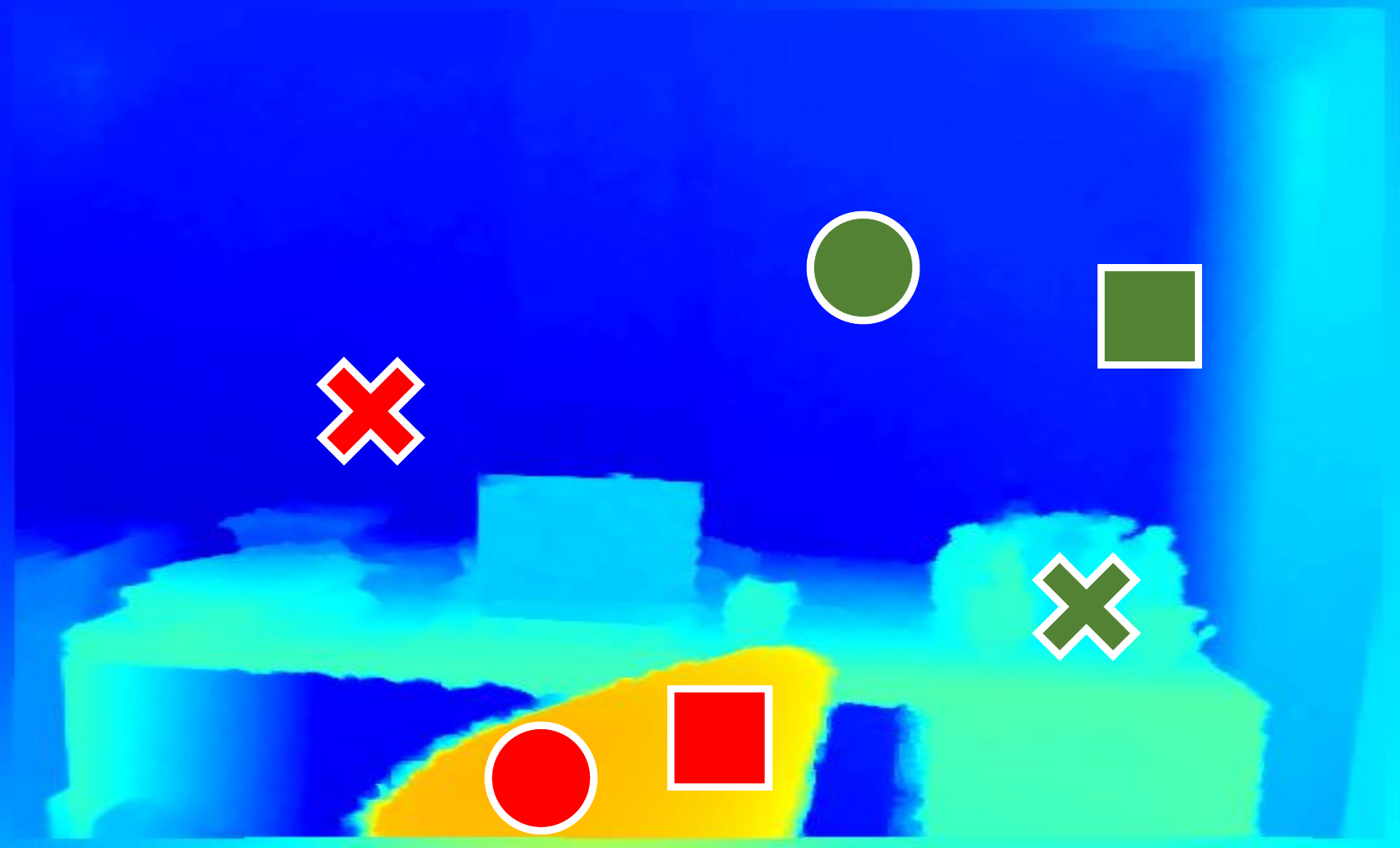}
}
\subfigure[Distance between SemSeg features w/ Depth labels]{
    \includegraphics[width=0.45\linewidth]{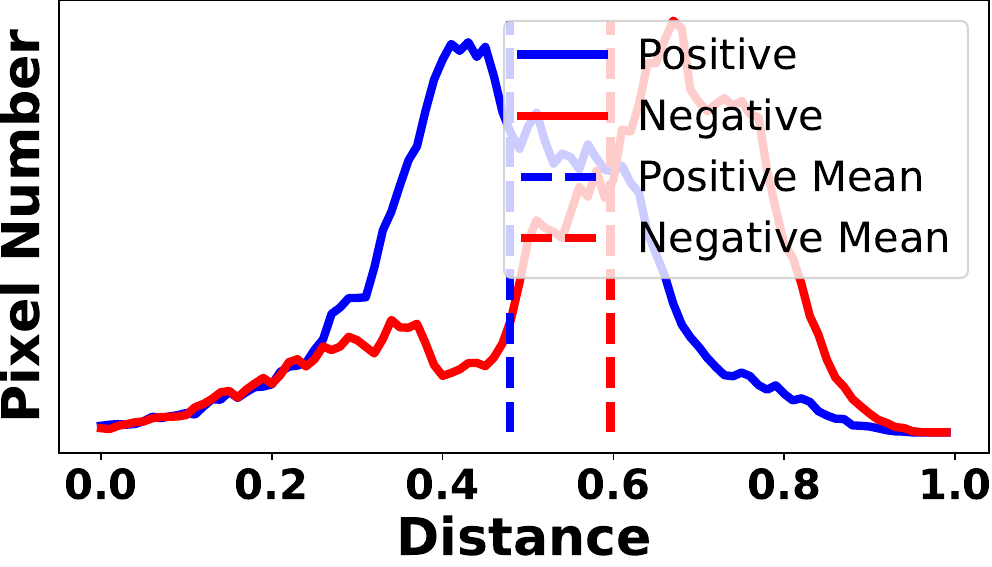}
}
\vspace{-10pt}
\caption{
The feature distance distributions of positive/negative pairs in triplets sampled from the same projected semantic feature maps while defined with different source task (\ie~SemSeg and Depth) labels show high cross-task consistency, which can be utilized to boost multi-task pixel-wise learning and predictions.
In \textbf{(a)} and  \textbf{(c)}, anchor pixels are denoted with $\square$, positive pixels with $\circ$, and negative pixels with $\times$. \textbf{(b)} and \textbf{(d)} illustrates the feature distributions of the positive/negative pixel pairs in the projection feature space.
}
\label{fig:consistency}
\vspace{-16pt}
\end{figure}

\par Multi-task dense prediction~\cite{kanakis2020reparameterizing,vandenhende2020mti,bruggemann2021exploring} offers an effective solution to this problem by jointly learning both task-sharing and task-specific representations. As the task-sharing representations can be obtained from a common backbone, it greatly facilitates the reduction of memory and computation overhead in both training and testing. However, simply using a shared backbone and several individual decoding heads for different tasks often causes a clear performance drop compared to a single-task dense prediction framework~\cite{xu2018pad}. Therefore, how to design a deep network structure that can effectively model cross-task interactions based on the complementarity of different tasks is critical for multi-task dense prediction. 

\par To achieve the above-mentioned objective, recent works on multi-task dense prediction mainly employ two paradigms. One is to refine final task-specific features with a multi-modal distillation module via carefully designed attention mechanisms to improve the final prediction~\cite{xu2018pad,vandenhende2020mti,bruggemann2021exploring}. The other is to learn and combine intra- or inter-task affinity maps to refine the task-specific features~\cite{zhang2019pattern,zhou2020pattern}. A common point for these two categories of methods is that they need to have extra expensive network computation modules for cross-modal or cross-task interactions, especially when the computation is performed in a multi-scale setting, which will significantly increase the training and inference cost.

In this work, in contrast to existing methods focusing on network structure improvement, we present a novel approach for the targeted problem based on learning cross-task contrastive consistency regularization upon dense pixel-wise features. The intuition of of utilizing contrastive consistency for modeling cross-task interactions is mainly threefold. First, as we learn the different tasks from the same input image data, the consistency inherently exists among the different tasks (see Fig.~\ref{fig:consistency}). Second, the features from different task decoders corresponding to the same semantic object categories should be more similar and consistent in the feature space than those with different object categories. Third, the extra network modules for explicit cross-task distillation in previous works inevitably increase the complexity of the multi-task model, bringing larger computation overhead in both the training and testing phases. However, a contrastive consistency optimization objective on multi-task features would only bring computation in the training stage, while the model size and the testing efficiency can be effectively improved. 

Based on these motivations, we develop a contrastive learning approach for multi-task dense prediction, which leverages cross-task consistency and applies contrastive regularization onto the features of different tasks. The feature-level multi-task contrastive optimization objective can guide the model to learn more effective task-specific features via absorbing complementary information from other tasks, without an explicit utilization of extra cross-task interaction network structures.
\begin{figure*}[!t]
    \centering
    \includegraphics[width=0.99\textwidth, height=0.359\textwidth]{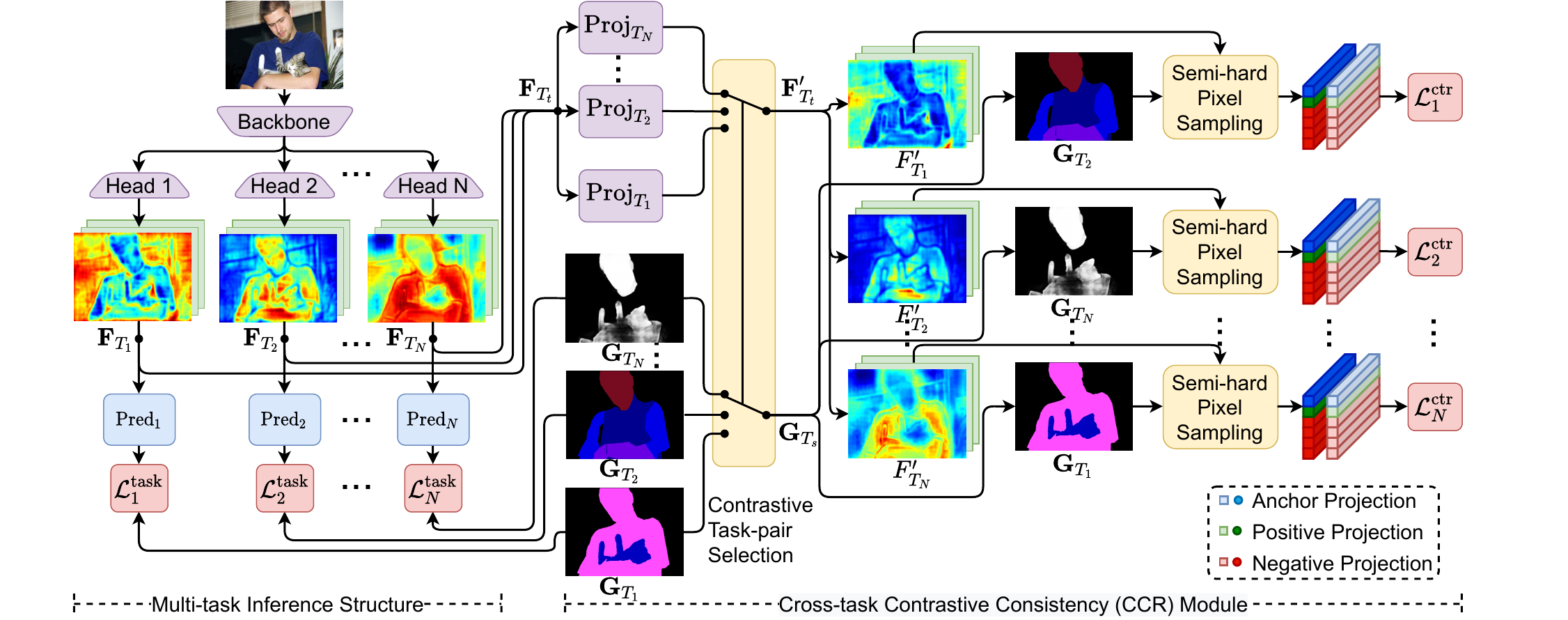}
    \vspace{-6pt}
    \caption{Overview of a multi-task dense prediction model with the proposed cross-task contrastive regularization learning method. It should be noted that a target task does not consider itself as an available source task. The proposed cross-task contrastive regularization module can be flexibly utilized by any existing multi-task dense prediction models with an encoder-decoder design. At each training iteration, a random source task is selected for every target task independently, whose ground-truth label $\mathbf{G}_{T_s}$ is used to guide the regularization applied on target-task feature map $\mathbf{F}_{T_t}$. A projector $\text{Proj}_{T_t}$ for every pair of target and source tasks produces a feature projection map $\mathbf{F}'_{T_t}$. A sampling strategy guided with positive/negative definition based on $\mathbf{G}_{T_s}$ is applied on projection maps, and the selected pixels are then used for the triplet loss. Best viewed in color.}
    \vspace{-15pt}
    \label{fig:frameworkoverview}
\end{figure*}
The proposed approach can also be flexibly generalized, in a plug and play fashion, to other multi-task dense prediction frameworks that may have different network architectures and objective functions for discrete or continuous dense prediction tasks.  

The main contribution of this paper is threefold:
\begin{itemize}
    \item We demonstrate pixel-level feature consistency exists among multiple distinct dense prediction tasks and introduce feature-wise contrastive consistency to guide the learning of discriminative multi-task dense features, and propose a novel cross-task contrastive learning strategy for the problem of joint multi-task dense predictions.

    \item 
    We implement an end-to-end multi-task contrastive regularization framework based on the feature consistency, and further design effective schemes including a generic criterion for positive/negative definition for both continuous and discrete tasks,
    shared feature projection, semi-hard pixel sampling, and contrastive task-pair selection to advance the framework. 
    
    \item Extensive experiments on two challenging datasets (\ie~NYUD-v2 and Pascal-Context) clearly demonstrate the effectiveness of the proposed cross-task contrastive regularization model for multi-task dense prediction, establishing new state-of-the-art performance. The results also verify that the proposed model can be generalized to different existing multi-task dense prediction frameworks to boost the performance and costs no additional computation in inference.
\end{itemize}

\section{Related Work}\label{sec:related work}
We review closely related works on multi-task dense prediction and contrastive learning for vision problems.
\subsection{Multi-task Dense Prediction}\label{subsec:MTL}
Recent methods on multi-task dense prediction~\cite{vasu2021instance,sun2021task,li2020knowledge, li2022Learning} can be roughly divided into two groups: the first group of methods~\cite{kendall2018multi,chen2018gradnorm,neven2017fast,sener2018multi,teichmann2018multinet,gao2019nddr,liu2019end} shares task-specific information in the encoding stage while the other group of methods shares the information in the decoding stage~\cite{xu2017multi,bruggemann2021exploring,zhang2019pattern,zhou2020pattern}. We focus on the review of the second group of methods since their setting is closer to ours.

One common practice among decoder-focused methods is to use features from other tasks to refine the final features or predictions via cross-task or cross-modal distillation~\cite{xu2018pad,vandenhende2020mti,bruggemann2021exploring}. More specifically, PAD-Net~\cite{xu2018pad} applies a multi-task distillation with a spatial attention mechanism to enhance task-specific predictions while at a fixed scale, leading to sub-optimal performance.
To alleviate this issue, MTI-Net~\cite{vandenhende2020mti} proposes to apply multi-modal distillation at different scales in a parallel way. Despite distillation at different scales brings diverse receptive fields, the information interaction crossing tasks is still restricted in a local context.
Thus, Bruggemann~\etal~\cite{bruggemann2021exploring} design an ATRC module to leverage both local and global contexts, based on not only task-specific features as in PAD-Net and MTI-Net but also task-specific predictions in a dynamic and adaptive manner. Another research line of decoder-focused models, for instance, PAP~\cite{zhang2019pattern} and PSD~\cite{zhou2020pattern}, aims to refine task-specific feature maps by mining local or global affinities on task-specific features. 

In contrast to these two groups of methods which design extra network modules to explicitly perform cross-task interaction and refine features during inference, our method directly regularizes the model learning to produce more effective multi-task features during training, using a novel multi-task contrastive learning approach. Thus, our model can simplify the model complexity, and can also be more flexibly applied in a plug and play way into other multi-task dense prediction frameworks.

\subsection{Contrastive Learning for Vision}\label{subsec:CTL}
Contrastive learning has been widely used to deal with various fundamental representation learning and its application to downstream single tasks problems~\cite{van2018representation,he2020momentum,chen2020improved,chen2020simple,chen2020big}. 
One challenging \emph{single} dense prediction task, \ie~semantic segmentation, has greatly benefited from the contrastive learning philosophy in its semantic representation learning~\cite{hu2021region,alonso2021semi,zhao2021contrastive,zhong2021pixel,chaitanya2020contrastive,van2021unsupervised}. 
For instance, RegionContrast~\cite{hu2021region} applies contrastive regularization on features to enhance the similarity between pixels corresponding to the same category, while Zhao~\etal~\cite{zhao2021contrastive} adopt contrastive learning as a pretraining strategy to tackle with performance drops when training data is insufficient.
Some other works leverage contrastive learning for semantic segmentation in a self-supervised~\cite{zhang2021looking,van2021unsupervised} or semi-supervised~\cite{chaitanya2020contrastive,zhong2021pixel,alonso2021semi} manner.

However, in contrast to these works considering learning only with a \emph{single} dense prediction task, our work aims at improving the comprehensive performance of multi-task dense predictions involving multiple distinct tasks via cross-task contrastive consistency.

\section{Contrastive Multi-Task Dense Prediction}\label{sec:method}

A framework overview of the proposed approach is shown in Fig.~\ref{fig:frameworkoverview}.
The proposed contrastive multi-task regularization model is constructed based on dense feature consistency, and is applied upon different task-specific feature maps from the multiple decoding heads, to boost their representation learning. It consists of several designed important components, including shared feature projectors, sampling strategies of spatial feature points of different tasks, the definition criterion of positive and negative samples, and the multi-task contrastive learning objective. In the following, we first discuss the cross-task feature consistency, and then elaborate the details of the proposed model. 

\begin{figure*}[!t]
    \centering
    \includegraphics[width=\textwidth]{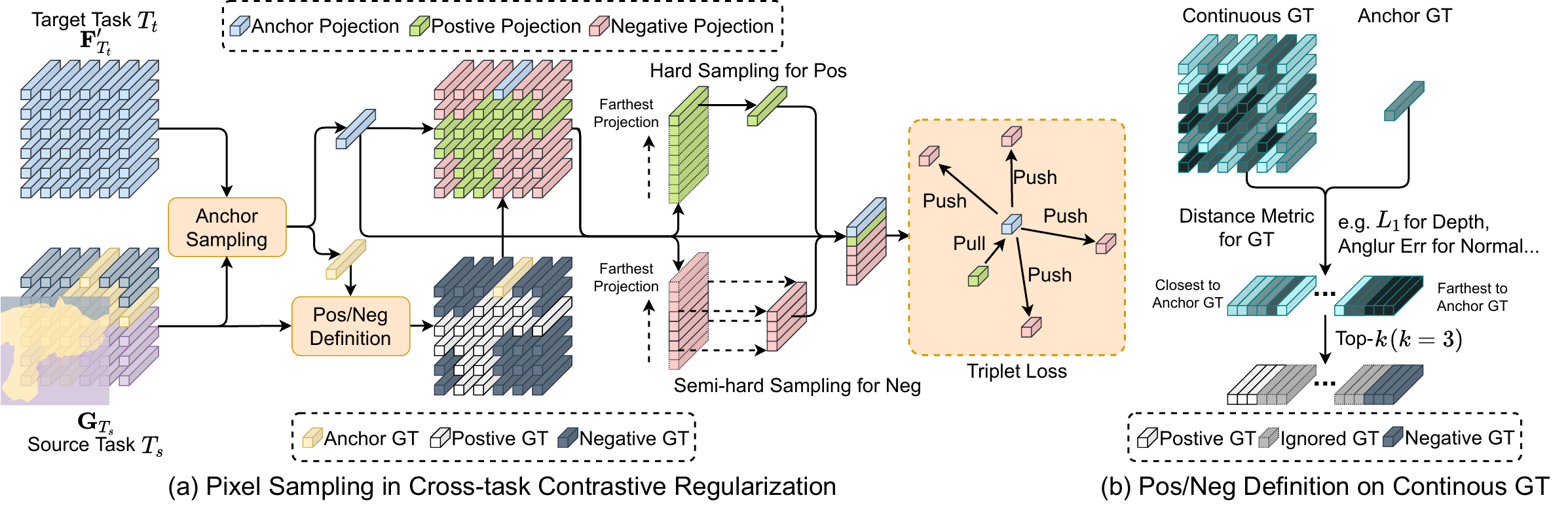}
    \vspace{-20pt}
    \caption{(a) shows a detailed illustration on the designed semi-hard pixel sampling process for the proposed Cross-task Contrastive Regularization (CCR) method. Anchor pixels are first sampled, whose corresponding labels are then compared with the labels of other pixels to determine whether the projection of a certain pixel should be regard as a positive or a negative sample. Hard sampling is applied on positive ones, while semi-hard sampling is applied for negative ones.
    (b) shows a generic positive/negative definition method for both continuous and discrete tasks. The definition for tasks with discrete ground-truths is illustrated in (a) while that for tasks with continuous ground-truths is elaborated in (b).}
    \label{fig:sampling}
    \vspace{-18pt}
\end{figure*}

\vspace{-3pt}
\subsection{Cross-Task Dense Feature Consistency}
\label{subsec:consistency}
From a basic principle of image-based representation learning, the pixels with the same ground-truth labels distribute more closely (\ie~with smaller distance) in the feature space compared to those with different ground-truth labels.
In the context of multi-task dense prediction, as the different tasks are learned from the same input image data, consistency inherently exists crossing the different tasks. The model performance of one task can thus benefit from utilizing the consistent information from other tasks.
To better illustrate such consistency, we also train the multi-task baseline model on NYUD-v2~\cite{silberman2012indoor}, and sample $10^5$ pixel triplets, each with one anchor pixel, one positive pixel and one negative pixel, from the feature maps that are used for final predictions. Detailed criterion of positive or negative sample definition and the training setting are elaborated in later sections.
The distance is calculated with a squared L2 between two normalized pixel features. As shown in Fig.~\ref{fig:consistency}, the feature distance distribution shows high consistency on different target tasks, \eg,~distances of positive pairs are consistently smaller than those of negative ones on different target tasks, and the distance distributions are also similar.
Based on such observations, we design a multi-task contrastive regularization model to utilize the cross-task consistency to enhance the multi-task representation learning.

\subsection{Multi-Task Contrastive Regularization}\label{subsec:learning obj}
We first introduce the definition criterion of pixel-wise positive and negative samples, and then present the details of the proposed pixel-wise feature contrastive regularization based on cross-task consistency for task-specific feature learning, and important schemes to boost the learning performance, including the shared feature projector, the semi-hard pixel sampling strategy, and the contrastive task-pair selection.

\vspace{-5pt}
\subsubsection{Positive/Negative Sample Definition}\label{subsec:posneg}
Dense prediction tasks can be divided into two types according to the continuity of the ground-truth label space, \ie~pixel-wise classification tasks with \textit{discrete} ground-truths, \eg,~semantic segmentation and human-parts parsing, and pixel-wise regression tasks with \textit{continuous} ground-truths, \eg,~depth and surface normal estimation. For the proposed multi-task dense contrastive learning objective, we introduce two general definition criteria for positive and negative samples for these two different types of tasks:

\noindent\textbf{(i)} {Tasks with discrete ground-truth labels.} It is relatively straightforward to define positive and negative pixel samples as the pixels with the same ground-truth labels can be considered as positives while those with different labels can be considered as negatives.
We simply apply this rule to determine positive and negative pixel samples for this group of dense prediction tasks. 

\noindent\textbf{(ii)} {Tasks with continuous ground-truth labels.} Different from those tasks with discrete ground-truth labels, there is no clear boundary to distinguish positive and negative samples for tasks with continuous ground-truth labels. Thus, for a certain anchor pixel that is randomly selected, we propose to utilize top-$k$ pixels with closer distance to the anchor pixel as positive samples, and top-$k$ pixels with farther distance to the anchor pixel as negative ones. Since different evaluation metrics for dense prediction tasks are usually based on pixel-wise comparison, the measurement of pixel-wise distances in the label space can be based on the evaluation metric for the corresponding task, \eg,~the L1 distance for depth estimation, and the angular error for surface normal estimation.

With the proposed definition criteria, we can ensure that our model can be flexibly generalized to any continuous or discrete dense prediction tasks.

\subsubsection{Pixel-wise~Feature~Contrastive~Regularization}
We now present the proposed cross-task contrastive regularization (CCR) for discriminative multi-task feature learning.
Given $N$ different dense predictions tasks, any task can be a source task $T_s$ ($s\in \{1,\dots,N\}$ or a target task $T_t$ ($t\in \{1,\dots,N\}$). To enforce the cross-task contrastive learning, $C_A$ anchor positions are first sampled from a target-task-specific feature map $\mathbf{F}_{T_t}$. The sampling in the basic CCR module considers a uniform sampling strategy. The ground-truth labels $\mathbf{G}_{T_s}$ of a source task $T_s$ are used to define the triplets based on the positive/negative sample definition presented in the previous section, and we generate a set of $C_A$ triplets $\{a_{T_s}^{c-}, a_{T_s}^c, a_{T_s}^{c+}\}_{c=1}^{C_A}$, where $a_{T_s}^{c+}$ and $a_{T_s}^{c-}$ are positive and negative pixels of an anchor pixel $a_{T_s}^{c}$, respectively.
Then, a cross-task contrastive regularization $\mathcal{R}^{\text{ctr}}_{ts}$ is defined as:
\begin{equation}
\setlength\abovedisplayskip{5pt}
\begin{aligned} 
 \mathcal{R}^{\text{ctr}}_{ts} = 
    \frac{1}{C_A} \sum_{c=1}^{C_A}\big[&D\big(\mathbf{F}'_{T_t}(a_{T_s}^c),  \mathbf{F}'_{T_t}(a_{T_s}^{c+})\big)
    - \\&
    D\big(\mathbf{F}'_{T_t}(a_{T_s}^c), \mathbf{F}'_{T_t}(a_{T_s}^{c-})\big) + m
    \big]_{+},
\end{aligned}
\end{equation}
where $D(\cdot)$ is a distance measurement function, and an L2 distance is utilized for simplicity. $\mathbf{F}'_{T_t}$ is a normalized feature map produced from a task-specific feature projector $\mathrm{Proj}_{T_t}(\cdot)$ as $\mathbf{F}'_{T_t} = \mathrm{Proj}_{T_t}(\mathbf{F}_{T_t})$. It can be represented as:
\begin{equation}
    \text{Proj}_{T_t}(\cdot) = \text{BN} \circ \text{Conv}_{1\times1} \circ \text{ReLU} \circ \text{Conv}_{1\times1}(\cdot)
\end{equation}
where $\mathrm{BN}$ is a batch normalization; $\mathrm{Conv_{1\times 1}}$ is a convolution with 1x1 kernel; $\mathrm{ReLU}$ is the ReLU activation function. $\mathbf{F}_{T_t}(\cdot)$ returns a feature vector given an input pixel location on the feature map. It should be noted that the ground-truth maps are downsampled to match the resolution of the projected feature map $\mathbf{F}_{T_t}$ for the contrastive loss computation. 
$m$ is a margin variable that is enforced between positive and negative feature points. If we utilize the contrastive regularization on every pair of a source task $T_s$ and a target task $T_t$, the overall multi-task contrastive regularization term writes:
\begin{equation}\label{equ:ctr_loss}
    \mathcal{L}^{\text{ctr}} = \sum_{t=1}^{N}{\sum_{s=1, s \neq t}^{N}{\mathcal{R}_{ts}^{\text{ctr}}}}\big(\mathbf{F}_{{T_t}}, \mathbf{G}_{T_s}, C_A, m\big).
\end{equation}
\par The computation of the above-proposed multi-task contrastive learning objective can be implemented as an end-to-end learnable module, \ie~the proposed Cross-task Contrastive Regularization (CCR) module as illustrated in~Fig.~\ref{fig:frameworkoverview}. A semi-hard pixel sampling scheme produces better sampled positive and negative feature points, and a contrastive task-pair selection determines which pair of tasks to be chosen for constructing an efficient regularization. We elaborate the details of these two components in the next parts. 
\subsubsection{Semi-hard Pixel Sampling}\label{subsec:sample}
Pixel features are considered as independent samples in our approach since the proposed multi-task contrastive regularization is applied in a pixel level. It is obvious that using all the pixels in every image requires very high computational overhead, and more importantly, not every pixel delivers critical information for the learning. Thus, we propose a pixel sampling strategy to address these issues. It follows several steps: \textbf{(i)} A certain amount of anchor pixels are uniformly sampled with a fixed ratio $\gamma$ to the total number of pixels on an image $\mathbf{I}\in \mathbb{R}^{H \times W}$ downsampled to the resolution of projection map $p_t$, where $H$ and $W$ indicate the height and width of the image, respectively. This process produces $C_A = \gamma HW$ anchors, \ie~$a_{T_s}^c \sim U(\mathbf{I})$. \textbf{(ii)} With the ground truth $\mathbf{G}_{T_s}$ from a source task $T_s$, for each anchor pixel $a_{T_s}^c$, all the remaining pixels in $\mathbf{I}$ unless ignored are divided into a positive pixel set $\mathbf{I}_{T_s}^{+}$ or a negative pixel set $\mathbf{I}_{T_s}^{-}$. \textbf{(iii)} To sample positive and negative pixels, the L2 distances between anchors and all the pixels in the projected feature space are calculated, resulting in a distance matrix $M \in \mathbb{R}^{C_A \times HW}$. To achieve a faster convergence, hard pixels should have higher priority to be sampled. \textbf{(iv)} For every anchor pixel $a_{T_s}^c$, the most distant positive pixel is selected as its only positive sample:
\begin{equation}
a_{T_s}^{c+} = \argmax_{a \in I_{T_s}^{+}}{D\left(\mathbf{F}'_{T_t}(a_{T_s}^c), \mathbf{F}'_{T_t}(a)\right)}. 
\end{equation}
\vspace{-2pt}
It is regarded as the hardest positive sample to be learned.
\textbf{(v)} We also use semi-hard mining~\cite{schroff2015facenet} to sample negative pixels to prevent the feature projector $\text{Proj}_{T_t}(\cdot)$ from collapsing.
For every anchor pixel $a_{T_s}^c$, a set of $C_\text{neg}$ negative pixels is selected, with each negative pixel $a_{T_s}^{c-}$ having the smallest distance with $a_{T_s}^c$ while still satisfying $\text{D}(\mathbf{F}'_{T_t}(a_{T_s}^c), \mathbf{F}'_{T_t}(a_{T_s}^{c-})) > \text{D}(\mathbf{F}'_{T_t}(a_{T_s}^c), \mathbf{F}'_{T_t}(a_{T_s}^{c+}))$. The proposed pixel sampling scheme is applied on each image independently to guarantee a large diversity of positive or negative pixel samples.

\subsubsection{Shared Feature Projector}
Feature projection is critically important in the contrastive learning. Since the contrastive regularization in our model is applied on different combinations of target and source tasks, in which each task provides feature map and ground-truth labels for the cross-task regularization. In other words, any task can be a source task or a target task, and the source task provides labels to regularize the feature map of the target task through the positive/negative sample selection. Considering multiple tasks are simultaneously learned, if we have all projectors unshared, the projector capacity for all the tasks is very large, which may increase the difficulty of learning. 
Thus, we design different sharing strategies for learning the projectors for the different source and target tasks. If all the target tasks use the same feature projector, we notate this projector design as `target-task shared ($T_{t}$-shared)'. If a target task uses the same projector for any paired source tasks, which provide distinct ground truth label for the cross-task regularization, we refer this projector as `source-task shared ($T_{s}$-shared)'. The proposed sharing strategy can help not only achieve the learning efficiency but also improve the performance. Detailed results and analysis can be found in the experiments.

\subsubsection{Contrastive Task-pair Selection}\label{subsec:tasksample}
Each cross-task constrastive regularization is constrained on a pair of tasks. When the number of tasks (\ie~$N$) is very large, we have $N\times(N-1)$ possible combinations, which brings tremendously large-scale contrastive computation in the learning process. To simplify and make a reasonable optimization space, we propose a contrastive task-pair selection strategy to address this issue. Specifically, in each iteration of the optimization, for each target task $T_t$ ($t\in \{1,...,N\}$), we randomly select one source task from the rest $N-1$ tasks. Since we perform random sampling, for each iteration, the same target task may pick up a different source task. After the whole training procedure, each target task can approximately obtain contrastive regularization from any source task in the whole task set. 

\subsection{Model Implementation Details}
\par\noindent\textbf{Overall Learning Objective}
The overall multi-task learning objective $\mathcal{L}_{overall}$ can be written as:

\vspace{-4pt}
\begin{equation}
    \label{equ:overall loss}
    \mathcal{L}_{overall} = \sum_{i=1}^{N} \lambda_i^{\text{task}} \mathcal{L}_i^{\text{task}} + \lambda^{\text{ctr}} \mathcal{L}^{\text{ctr}},
\end{equation}
\vspace{-3pt}
where $\mathcal{L}_i^\text{task}$ is a task-specific optimization loss for task $T_i$ (\ie~a classification or a regression loss), and $\lambda_i^{\text{task}}$ denotes a loss weight for each task.  $\lambda^\text{ctr}$ is the loss weight for all the cross-task contrastive losses. The $\lambda^\text{ctr}$ is linearly ramped up in the first several epochs to stabilize the training.
\vspace{2pt}
\par\noindent\textbf{Model Inference} In the inference stage, the cross-task contrastive learning module can be removed from the whole multi-task framework for the inference of different tasks, which is also a significant advantage compared to previous multi-task dense prediction frameworks that design extra expensive distillation modules for the cross-task interaction, and also needs to involve the modules in the inference stage.

\section{Experiments}\label{sec:exp}
\subsection{Experimental Setup}
\par\noindent\textbf{Datasets}
The experiments are extensively conducted on two widely used multi-task dense prediction datasets.
One is NYUD-v2~\cite{silberman2012indoor} which contains 1,449 RGBD indoor scene images with annotations for tasks of semantic segmentation (SemSeg), depth estimation (Depth), and surface normal estimation (Normal),
with 795 images for training and 654 images for testing. 
The other one is Pascal-Context~\cite{everingham2010pascal} which has 4,998 training and 5,105 testing images labeled for tasks of semantic segmentation (SemSeg), human-parts parsing (Parsing), saliency estimation (Saliency), surface normal estimation (Normal), and edge detection (Edge).

\begin{table}[!t]
\centering
\resizebox{0.95\linewidth}{!}{
\begin{tabular}{lcccc}
\toprule[1.2pt]
Model & SemSeg$\uparrow$ & Depth$\downarrow$ & Normal$\downarrow$ & $\Delta_m(\%)\uparrow$\\
\midrule[0.8pt]
ST Baseline & 39.803 & 0.617 & 19.896 & -\\
\midrule[0.8pt]
MT Baseline & 38.901 & 0.615 & 20.712 & -2.025\\
CCR-Basic & 39.499 & 0.609 & 20.305 & 0.184\\
% CCR w/ proj & 40.026 & 0.604 & 20.338\\
+ Proj & 39.956 & 0.603 & 20.297 & 0.635\\
+ Proj + SS & 40.779 & 0.595 & 20.240 & 1.203\\
+ Proj + SS + CTS & \textbf{41.275} & \textbf{0.592} & \textbf{20.178} & \textbf{2.101}\\
\bottomrule[1.2pt]
\end{tabular}
}
\vspace{-5pt}
\caption{Overall ablation study to show the effectiveness of the proposed cross-task contrastive regularization (CCR).}
\label{tab:proj_and_ctp}
\vspace{-6pt}
\end{table}

\begin{table}[!t]
\centering
\resizebox{0.95\linewidth}{!}{
\begin{tabular}{lcccc}
\toprule[1.2pt]
Model & SemSeg$\uparrow$ & Depth$\downarrow$ & Normal$\downarrow$\\
\midrule[0.8pt]
PAD-NET~\cite{xu2018pad} & 35.406 & 0.670 & 21.991 \\
\textbf{PAD-NET + CCR} & \textbf{36.925} & \textbf{0.654} & \textbf{21.139} \\
\midrule[0.8pt]
MTI-NET~\cite{vandenhende2020mti} & 37.456 & 0.626 & 21.031 \\
\textbf{MTI-NET + CCR} & \textbf{39.192} & \textbf{0.607} & \textbf{20.826} \\
\midrule[0.8pt]
InvPT~\cite{invpt2022} & 52.840 & 0.514 & 18.872\\
\textbf{InvPT + CCR} & \textbf{53.799} & \textbf{0.508} & \textbf{18.670}\\
\bottomrule[1.2pt]
\end{tabular}}
\vspace{-5pt}
\caption{Performance improvements with the proposed CCR applied on different best performing models on NYUD-v2.}
\label{tab:multi_baseline}
\vspace{-12pt}
\end{table}

\vspace{2pt}
\par\noindent\textbf{Evaluation Metrics} 
Following~previous~works~\cite{vandenhende2020mti},
SemSeg and Parsing tasks are
are evaluated with mean Intersection over Union (mIoU), 
Depth
with root mean square error (rmse), 
Normal 
with mean angular error (mErr), 
Salience
with maximum $F_1$ score (maxF1), and Edge
with optimal-dataset-scale F-measure (odsF). The overall multi-task performance introduced in~\cite{maninis2019attentive} is measured with an average of per-task performance differences {w.r.t.} a corresponding single-task baseline trained separately.

\par\noindent\textbf{Training Settings}
Similar to MTI-Net~\cite{vandenhende2020mti}, our ablation studies are extensively conducted on NYUD-v2 and the HRNet18~\cite{wang2020deep} is used as the model backbone which is pretrained on ImageNet~\cite{deng2009imagenet}. We train each model using Adam optimizer with a batch size of 4 on 2 GPUs (\ie~NVIDIA RTX 3090) for the NYUD-v2 dataset, and a batch size of 6 on 6 GPUs for the PASCAL-Context dataset. The base learning rate, momentum, and weight decay are set to 2e-4, 0.9, and 1e-4, respectively.  
The learning rate is linearly warmed up for 1 epoch. 
The margin $m$, sampling ratio $\gamma$, top-$k$ factor $k$, and contrastive loss weight $\lambda^{\text{con}}$ are by default set to 0.2, 0.01, 128, and 1.0 respectively in the evaluation experiments. The number of negative samples in one triplet is 16.

\subsection{Model Analysis}
\vspace{2pt}
\par\noindent\textbf{Model Baselines and Variants} Different baselines and model variants are considered for the evaluation: (i) \textbf{`ST Baseline'} is a very strong single-task baseline model using a shared FPN encoder~\cite{lin2017feature} with the HRNet18 backbone and a SemanticFPN decoder~\cite{kirillov2019panoptic}, which is similar to the baseline widely used in existing state-of-the-art models~\citep{vandenhende2020mti}, while with better performance.
(ii) \textbf{`MT Baseline'} indicates a multi-task baseline, which uses the same encoder architecture as the `ST Baseline', while it has task-specific decoders for the different tasks which are jointly optimized with the encoder under the multi-task learning setting. 
(iii) \textbf{`CCR-Basic'} directly applies the proposed
contrastive regularization on the final multi-task feature maps, based on the `MT Baseline'. (iv) \textbf{`+ Proj'} indicates applying a shared feature projector on the multi-task features before we feed them into the CCR module. (v) \textbf{`+ SS'} means using the designed semi-hard pixel sampling. (vi) \textbf{`+ CTS'} denotes using the proposed contrastive task-pair selection strategy.

\begin{table}[!t]
\centering
\resizebox{1.0\linewidth}{!}{
\begin{tabular}{lccccc}
\toprule[1.2pt]
Model & $T_{t}$-shared & $T_{s}$-shared & SemSeg$\uparrow$ & Depth$\downarrow$ & Normal$\downarrow$\\
\midrule[0.8pt]
MT Baseline & & & 38.901 & 0.615 & 20.712 \\
CCR & & & 40.414 & 0.598 & 20.219\\
CCR & $\checkmark$ &  & 40.273 & 0.604 & 20.318\\
CCR & $\checkmark$ & $\checkmark$ & 40.574 & 0.600 & 20.299\\
CCR &  & $\checkmark$ & \textbf{41.275} & \textbf{0.592} & \textbf{20.178}\\
\bottomrule[1.2pt]
\end{tabular}
}
\vspace{-5pt}
\caption{Ablation study on feature projector designs. $T_{t}$/$T_{s}$-shared means projectors are shared among different target/source tasks. SS and CTS are included in all CCR models. }
\label{tab:proj}
\vspace{-9pt}
\end{table}

\begin{figure}[!t]
    \centering
    \includegraphics[width=1.01\linewidth]{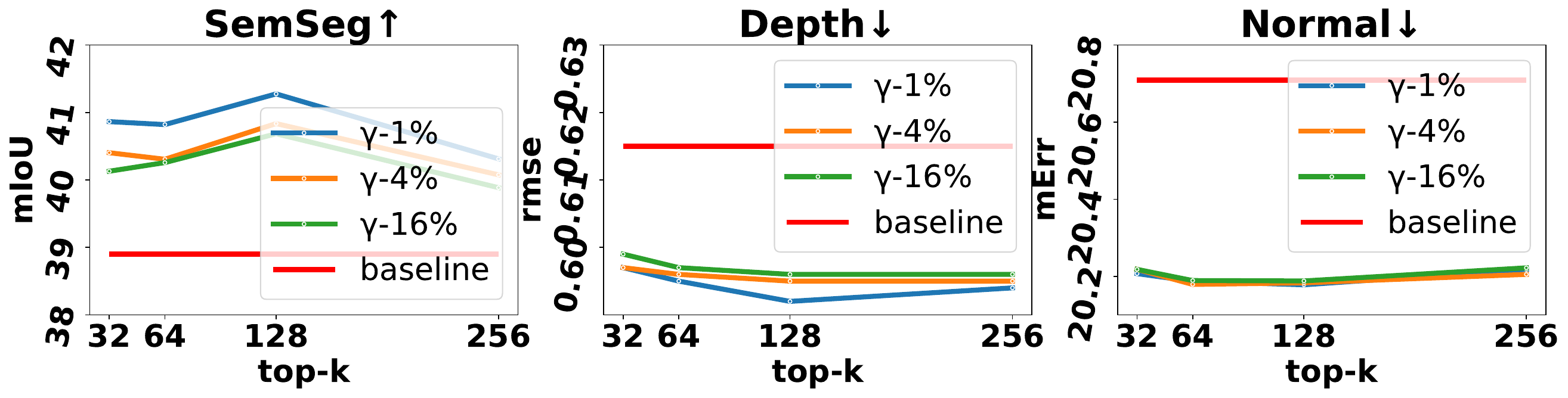}
    \vspace{-19pt}
    \caption{Investigation on top-$k$ and sampling ratio $\gamma$. It can be observed from the performance variance that the model is not very sensitive to $k$ and $\gamma$, while $k$ has a greater influence on model performance than $\gamma$ does. The best performance is achieved when $k$ is set to 128.
    }
    \label{fig:sample_k_ablation} 
    \vspace{-16pt}
\end{figure}

\vspace{2pt}
\par\noindent\textbf{Overall Ablation Study}
To validate the effectiveness of our approach, an overall ablation study is shown in Table~\ref{tab:proj_and_ctp}. 
Our basic CCR module (\ie~CCR-Basic) can bring notable performance gain over the baseline model (\ie~MT Baseline) on all the tasks, especially on SemSeg and Depth. The addition of the proposed shared projector strategy (\ie~Proj) further improves the performance on all the tasks.
More experiments about the projector design are discussed later in this section. The semi-hard pixel sampling (\ie~SS) can also clearly boost the model's multi-task performance. Further more, it can be seen that the the contrastive task-pair selection strategy (\ie~CTS) which is originally designed to reduce the memory consumption, can also contribute to the model's performance. This may be due to that choosing one source task for each target task instead of pairing all the other tasks during each iteration can alleviate the gradient conflict issue on learning the shared source task projector.

\par\noindent\textbf{Study of Projector Designs} 
We study different projector designs and show the results in Table~\ref{tab:proj}. 
It can be observed that how projectors are shared among different constrastive task pairs can greatly affect the performance. We find that sharing projectors among source tasks achieves the best performance, which indicates that the cross-task contrastive regularization should be applied in a joint feature projection space regardless of which source task is used for the definition of positive and negative pairs, while sharing projectors among different target tasks leads to less optimal performance. This is because that the feature distributions of different target tasks vary largely, and we need different projectors for the distinct target tasks to learn the joint feature projection space for the multi-task contrastive learning.

\begin{table}[!t]
\centering
\resizebox{0.9\linewidth}{!}{
\begin{tabular}{lcccc}
\toprule[1.2pt]
Model & SemSeg$\uparrow$ & Depth$\downarrow$ & Normal$\downarrow$\\
\midrule[0.8pt]
PAP~\cite{zhang2019pattern} & 36.72 & 0.618 & 20.82\\
PSD~\cite{zhou2020pattern} & 36.69 & 0.625 & 20.87\\
ATRC~\cite{bruggemann2021exploring} & 46.33 & 0.536 & 20.18\\
InvPT~\cite{invpt2022} & 52.84 & 0.514 & 18.87\\
\midrule[0.8pt]
\textbf{CCR (Ours)} & \textbf{53.80} & \textbf{0.508} & \textbf{18.67}\\
\bottomrule[1.2pt]
\end{tabular}}
\vspace{-8pt}
\caption{State-of-the-art comparison on NYUD-v2. 
}
\label{tab:sota_nyud_3task}
\vspace{-10pt}
\end{table}

\begin{table}[!t]
\centering
\resizebox{1.0\linewidth}{!}{
\begin{tabular}{lccccc}
\toprule[1.5pt]
    Model & SemSeg$\uparrow$ & Parsing$\uparrow$ & Saliency$\uparrow$ & Normal$\downarrow$ & Edge$\uparrow$\\
\midrule[1pt]
PAD-NET~\cite{xu2018pad} & 53.60 & 59.60 & 65.80 & 15.30 & 72.50\\
MTI-NET~\cite{vandenhende2020mti} & 61.70 & 60.18 & 84.78 & 14.23 & 70.80\\
PSD~\cite{zhou2020pattern} & 61.70 & 60.18 & 84.78 & 14.23 & 70.80\\
ATRC~\cite{bruggemann2021exploring} & 67.67 & 62.93 & 82.29 & 14.24 & 72.42\\
InvPT~\cite{invpt2022} & 79.03 & 67.61 & \textbf{84.81} & 14.15 & 73.00\\
\midrule[1pt]
\textbf{CCR (Ours)} & \textbf{80.23} & \textbf{68.40} & {84.36} & \textbf{13.93} & \textbf{73.07}\\
\bottomrule[1.5pt]
\end{tabular}}
\vspace{-8pt}
\caption{State-of-the-art comparison on Pascal-Context. 
}
\label{tab:sota_pascal}
\vspace{-10pt}
\end{table}

\begin{figure}[!t]
\centering
\subfigure[Depth feature distance distribution  w/ Normal labels]{
    \includegraphics[width=0.98\linewidth]{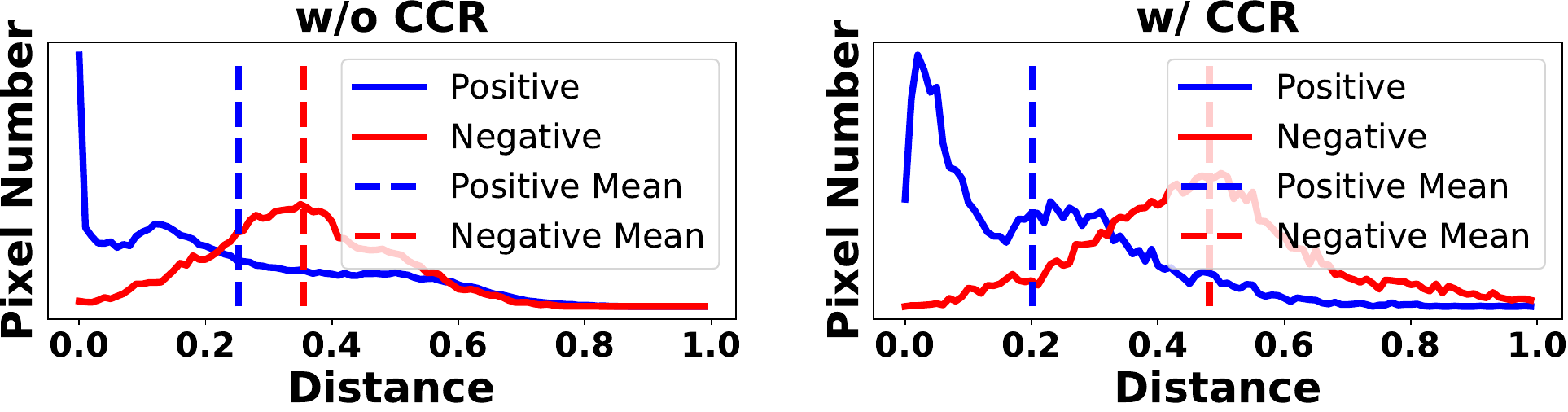}
    % \vksip -50pt
}
\vskip -5pt
\subfigure[SemSeg feature distance distribution w/ Depth labels]{
    \includegraphics[width=0.98\linewidth]{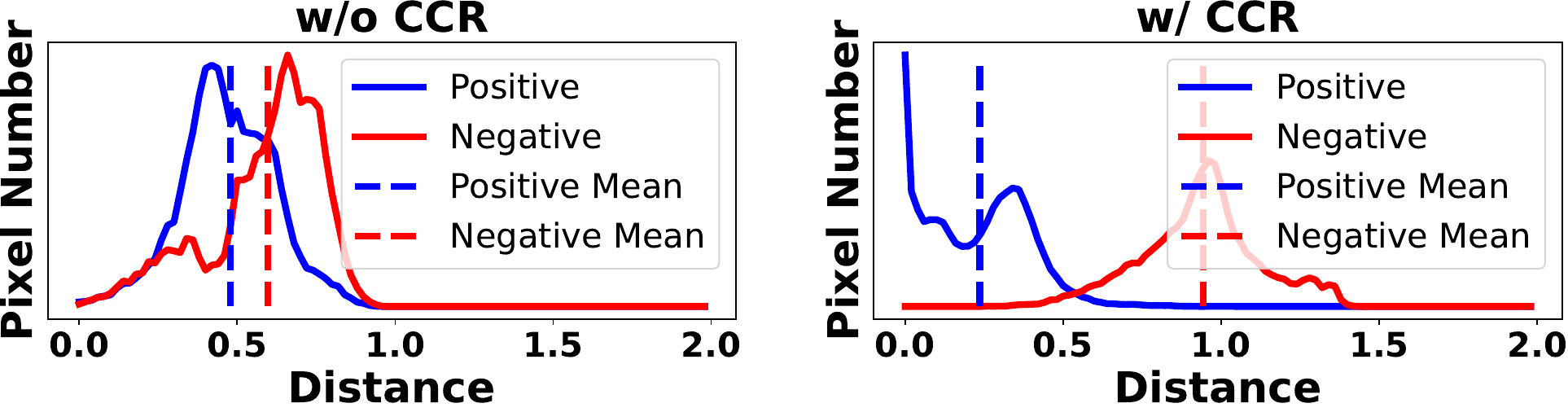}
}
\vspace{-12pt}
\caption{Feature distance distributions of feature maps of different tasks, \ie~\textbf{(a)} Depth and \textbf{(b)} SemSeg, based on sampled positive/negative pixel pairs defined using the 
different source task labels, \ie~\textbf{(a)} Normal and \textbf{(b)} Depth.
Distance in the projection feature space is measured with normalized $l2$ distance. After using the proposed CCR, the average distance of the positive and negative features are significantly pushed apart, achieving more discriminative features.
}
\label{fig:consistency_w/_ccr}
\vspace{-16pt}
\end{figure}

\par\noindent\textbf{Study of Top-$k$ Factor and Sampling Ratio $\gamma$} Since there are no discriminative boundaries between positive and negative samples based on continuous ground-truth labels, we introduce top-$k$-based method to define positive/negative samples as illustrated in the method section.  
If $k$ is too small, little contrastive constraint will be provided as sampled positive and negative pixels can be easily discriminated with the triplet contrastive loss, while a very large $k$ also brings ambiguity in the sample definition, and even causes intersection between positive and negative pixel samples, resulting in a learning corruption of the model.
Figure~\ref{fig:sample_k_ablation} shows an evaluation of the influence of the factor $k$ on the performance of different tasks.
The best multi-task performance is achieved when $k$ is set to 128, 
and 
the performance gets consistently improved as the sampling ratio
$\gamma$ increases, while the model performance is not very sensitive to the $k$ and $\gamma$.

\par\noindent\textbf{Study of Improvements on Different Baseline Models}
A simple yet effective regularization approach should be independent of the baseline models utilized. 
In Table~\ref{tab:multi_baseline}, we demonstrate the performance of our approach when applied upon the most advanced multi-task dense predictions models, including PAD-Net~\cite{xu2018pad}, MTI-Net~\cite{vandenhende2020mti}, and InvPT~\cite{invpt2022}.
It is clear that our approach can effectively boost the performances of all these baseline models, demonstrating the generalization ability of our model for multi-task dense prediction.

\par\noindent\textbf{Qualitative Study of the Effect on Feature Distributions}
To study the effect of the proposed CCR on the feature learning, 
we show the distance distributions of the pixel-wise feature triplets in the projection feature space, for both the baseline and our model in Fig~\ref{fig:consistency_w/_ccr}.
It can be observed that, with the proposed CCR, the pixel-wise features of different labels can be further pushed apart in the feature space, verifying our motivation of using the cross-task contrastive consistency for discriminative multi-task feature learning.

\begin{figure}[!t]
\centering
    \includegraphics[width=0.99\linewidth]{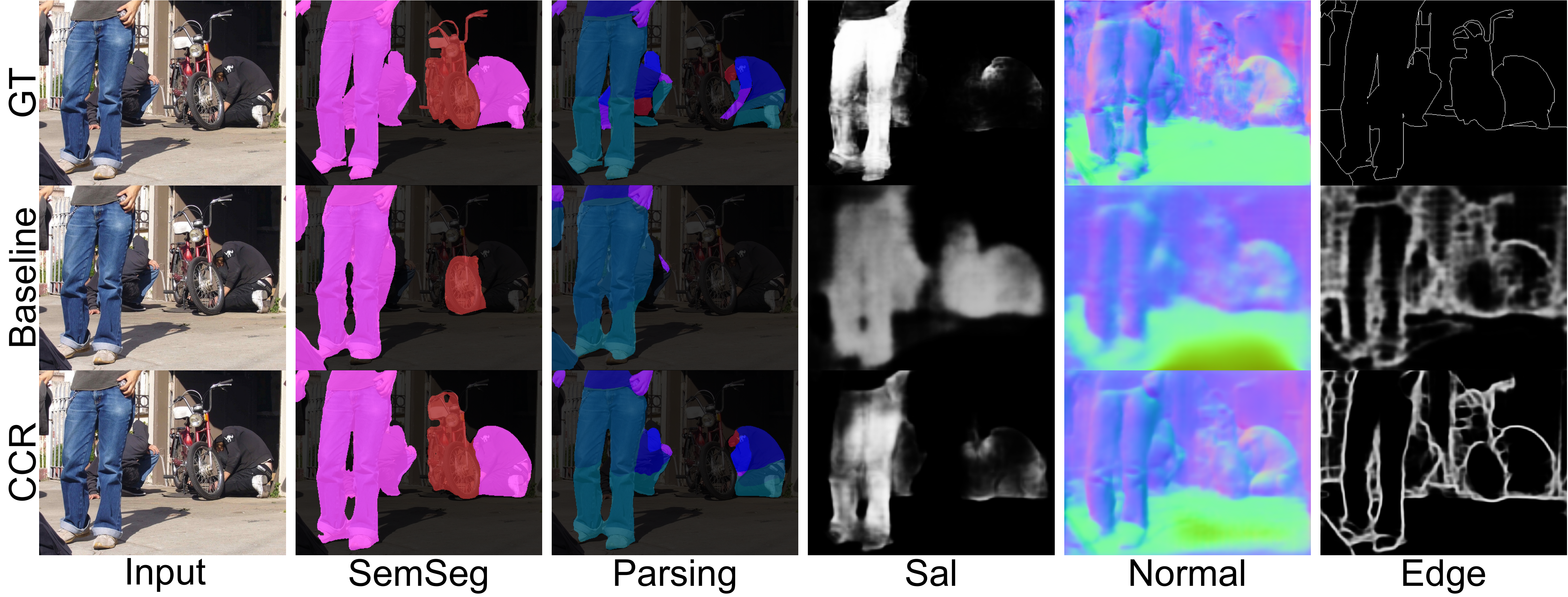}
    \vspace{-10pt}
    \caption{Qualitative comparison with the results of the baseline model and the ground-truths on Pascal-Context. CCR yields more accurate predictions on all the tasks.}
    \vspace{-16pt}
\label{fig:results_pascal}
\end{figure}

\subsection{State-of-the-art Comparison}
We compare the proposed contrastive multi-task learning approach with the single-task baseline, the multi-task baseline, and several best performing state-of-the-art multi-task dense prediction models in the literature, including PAD-Net~\cite{xu2018pad}, PAP~\cite{vandenhende2020mti}, MTI-Net~\cite{vandenhende2020mti}, PSD~\cite{zhou2020pattern}, and ATRC~\cite{bruggemann2021exploring}. The performance comparison is shown in Table~\ref{tab:sota_nyud_3task} and Table~\ref{tab:sota_pascal}. It can be seen that our approach applied on the strong multi-task baseline achieves the highest performance on the different tasks and on both the NYUD-v2 and Pascal-Context datasets, demonstrating the superiority of the proposed approach. 
Some qualitative comparison with the baseline model is also shown in Fig.~\ref{fig:results_pascal}.

\section{Conclusions}
We presented the proposed cross-task contrastive learning model, a novel regularization for the multi-task dense prediction problem based on cross-task contrastive consistency on task-specific features, and also further introduced several important components designed for the proposed contrastive model, including 
an effective pixel sampling strategy, a generic positive/negative definition criterion for both continuous and discrete tasks, shared feature projection scheme, and contrastive task-pair selection to reduce overhead. Extensive experiments on NYUD-v2 and PASCAL-Context clearly verified the effectiveness of the proposed approach.

\section*{Acknowledgements}
This research was partially supported by Early Career Scheme of the Research Grants Council (RGC) of the Hong Kong SAR under grant No. 26202321, and HKUST Startup Fund No. R9253.

\bibliography{aaai23}

\end{document}